\documentclass[letterpaper]{article} % DO NOT CHANGE THIS
\usepackage{aaai2026}  % DO NOT CHANGE THIS
\usepackage{times}  % DO NOT CHANGE THIS
\usepackage{helvet}  % DO NOT CHANGE THIS
\usepackage{courier}  % DO NOT CHANGE THIS
\usepackage[hyphens]{url}  % DO NOT CHANGE THIS
\usepackage{graphicx} % DO NOT CHANGE THIS
\usepackage{natbib}  % DO NOT CHANGE THIS AND DO NOT ADD ANY OPTIONS TO IT
\usepackage{caption} % DO NOT CHANGE THIS AND DO NOT ADD ANY OPTIONS TO IT
\usepackage{algorithm}
\usepackage{algorithmic}
\usepackage{import}

\usepackage{amsmath}  % new add

\usepackage{newfloat}
\usepackage{listings}

\usepackage{amssymb}  % new add
\usepackage{cleveref} % new add
\usepackage{booktabs} % new add
\usepackage{adjustbox}
\usepackage{multirow}
\usepackage{xcolor}
\usepackage[table]{xcolor}
\usepackage{pifont} % 添加 pifont 包以支持 \cmark 和 \xmark
\usepackage{booktabs} % 用于表格美化
\usepackage{enumitem}

\definecolor{backblue}{RGB}{210, 230, 250}

\definecolor{backred}{RGB}{255, 223, 223}

\definecolor{backgreen}{RGB}{220,244,229}

\definecolor{back_deepblue}{RGB}{180, 210, 240}

\definecolor{back_deepred}{RGB}{255, 200, 200}

\definecolor{back_deepgreen}{RGB}{190, 230, 210}

\definecolor{mygray}{gray}{0.95}

\definecolor{greentable3}{rgb}{0,0.5,0}

\newcommand{\cmark}{\ding{51}} % \ding{51} 是checkmark
\newcommand{\xmark}{\ding{55}} % \ding{55} 是xmark

\usepackage[most,skins,theorems]{tcolorbox}
\tcbset{
  takeawaysbox/.style={
    title=Takeaways,
    colback=lightblue!80,
    colframe=black,
    fonttitle=\bfseries\small,
    coltitle=white,
    colbacktitle=black,
    enhanced,
    attach boxed title to top left={xshift=2.5mm,yshift=-2.5mm},
    boxed title style={rounded corners, size=small, colframe=black, colback=black},
    width=\linewidth,
    arc=3.5mm
  }
}
\definecolor{lightblue}{RGB}{220,235,250}

\usepackage{newfloat}
\usepackage{listings}
\DeclareCaptionStyle{ruled}{labelfont=normalfont,labelsep=colon,strut=off} % DO NOT CHANGE THIS
\floatstyle{ruled}
\newfloat{listing}{tb}{lst}{}
\floatname{listing}{Listing}
\title{AStar: Boosting Multimodal Reasoning with Automated Structured Thinking}
\author {
    Jinyang Wu\textsuperscript{\rm 1},
    Mingkuan Feng\textsuperscript{\rm 1},
    Guocheng Zhai\textsuperscript{\rm 1},
    Shuai Zhang\textsuperscript{\rm 1,$^*$},
    Zheng Lian\textsuperscript{\rm 2},
    Fangrui Lv\textsuperscript{\rm 1}
    Pengpeng Shao\textsuperscript{\rm 1},
    Ruihan Jin\textsuperscript{\rm 1},
    Zhengqi Wen\textsuperscript{\rm 1},
    Jianhua Tao\textsuperscript{\rm 1,\thanks{Corresponding Author.}}
}

\affiliations {
    \textsuperscript{\rm 1}Department of Automation, BNRist, Tsinghua University\\
    \textsuperscript{\rm 2}Institute of Automation, Chinese Academy of Sciences\\
    wu-jy23@mails.tsinghua.edu.cn
}
\usepackage{bibentry}
\begin{document}

\maketitle

\begin{abstract}
Multimodal large language models excel across diverse domains but struggle with complex visual reasoning tasks. To enhance their reasoning capabilities, current approaches typically rely on explicit search or post-training techniques. However, search-based methods suffer from computational inefficiency due to extensive solution space exploration, while post-training methods demand substantial data, computational resources, and often exhibit training instability. To address these challenges, we propose \textbf{AStar}, a training-free, \textbf{A}utomatic \textbf{S}tructured \textbf{t}hinking paradigm for multimod\textbf{a}l \textbf{r}easoning. Specifically, we introduce novel ``thought cards'', a lightweight library of high-level reasoning patterns abstracted from prior samples. For each test problem, AStar adaptively retrieves the optimal thought cards and seamlessly integrates these external explicit guidelines with the model's internal implicit reasoning capabilities. Compared to previous methods, AStar eliminates computationally expensive explicit search and avoids additional complex post-training processes, enabling a more efficient reasoning approach. Extensive experiments demonstrate that our framework achieves 53.9\% accuracy on MathVerse (surpassing GPT-4o's 50.2\%) and 32.7\% on MathVision (outperforming GPT-4o's 30.4\%). Further analysis reveals the remarkable transferability of our method: thought cards generated from mathematical reasoning can also be applied to other reasoning tasks, even benefiting general visual perception and understanding. AStar serves as a plug-and-play test-time inference method, compatible with other post-training techniques, providing an important complement to existing multimodal reasoning approaches.

\end{abstract}

\section{Introduction}\label{sec1}

Multimodal Large Language Models (MLLMs) have demonstrated impressive capabilities across diverse tasks and domains~\citep{wang2024exploring,openai2024gpt4o,qiao2024we}, yet they struggle with complex visual reasoning tasks that require processing multimodal information with sophisticated problem-solving strategies~\citep{zhang2024multimodal,wang2024measuring}. Recent advances in System 2 slow-thinking reasoning models like OpenAI-o1~\citep{o1}, DeepSeek-R1~\citep{guo2025deepseek}, and Kimi-K1.5~\citep{team2025kimi} have inspired growing interest in incorporating structured long Chain-of-Thought (CoT)~\citep{wei2022chain} thinking into MLLMs~\citep{xu2024llava,dong2024insight,du2025virgo}. These approaches address the limitations of conventional MLLMs that often rely on simple ``direct prediction'' modes due to the scarcity of high-quality long-chain reasoning data~\citep{xu2024llava,luo2025ursa}. Current methods can be divided into two primary categories: (i) \textit{explicit search methods}~\citep{dong2024progressive,yao2024mulberry}: leverage algorithms like beam search or Monte Carlo Tree Search (MCTS) with specialized reward models to guide solution path exploration; (ii) \textit{post-training methods}~\citep{wang2024exploring,wang2025multimodal}: develop structured long CoT reasoning capabilities through common techniques such as Supervised Fine-Tuning (SFT)~\citep{zhang2024multimodal,luo2025ursa} or Reinforcement Learning (RL) like Proximal Policy Optimization (PPO)~\citep{schulman2017proximal} and Group Relative Policy Optimization (GRPO)~\citep{guo2025deepseek}.

\begin{figure}
    \centering
    \includegraphics[width=0.95\linewidth]{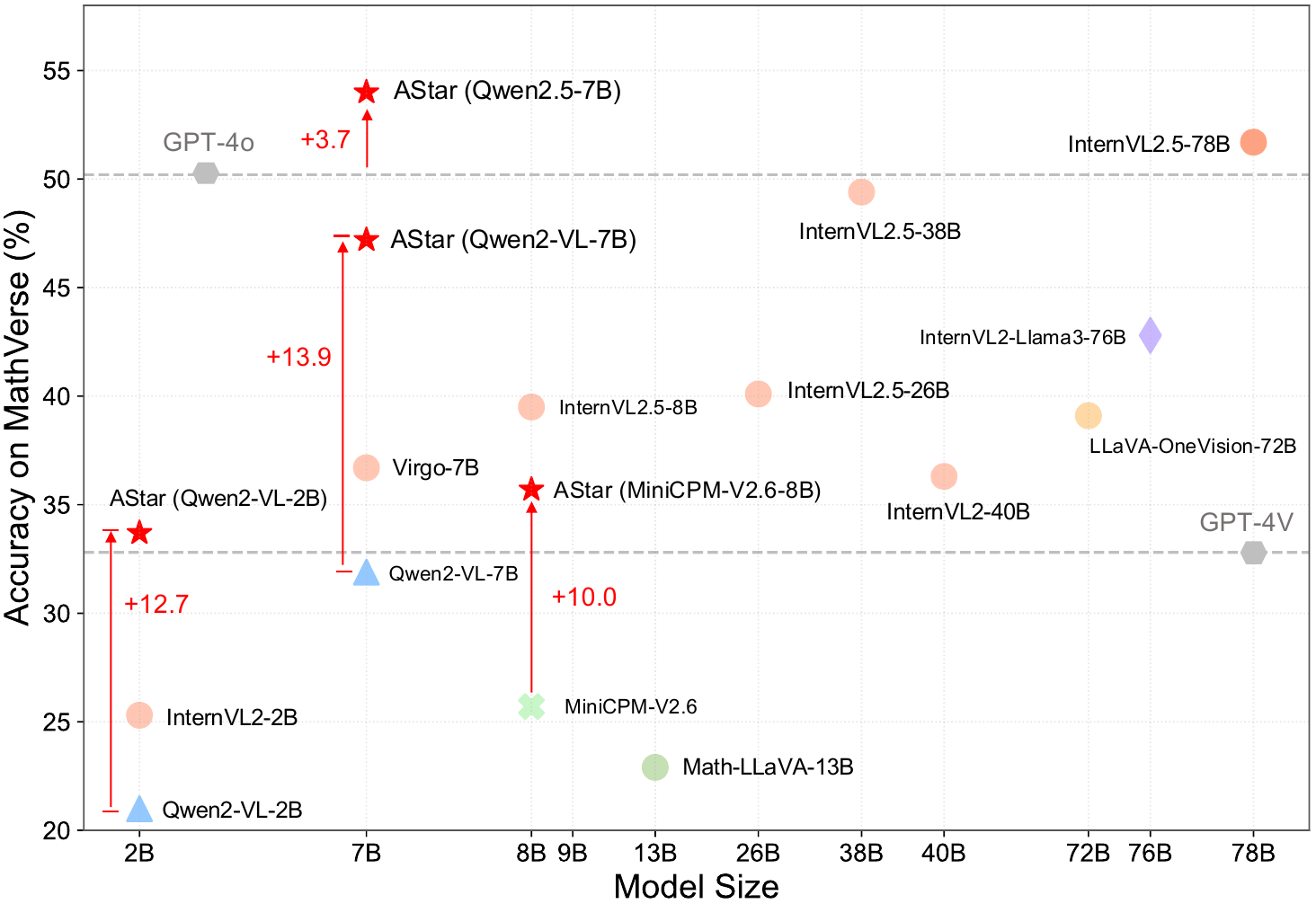}
    \caption{\textbf{Evaluation results on MathVerse.} AStar makes 7B models competent problem-solvers, surpassing GPT-4o. Our approach demonstrates consistent effectiveness across multiple model architectures and scales.}
    \label{Figure1}
\end{figure}

\begin{figure*}[ht!]
\begin{center}
\centerline{\includegraphics[width=0.96\textwidth]{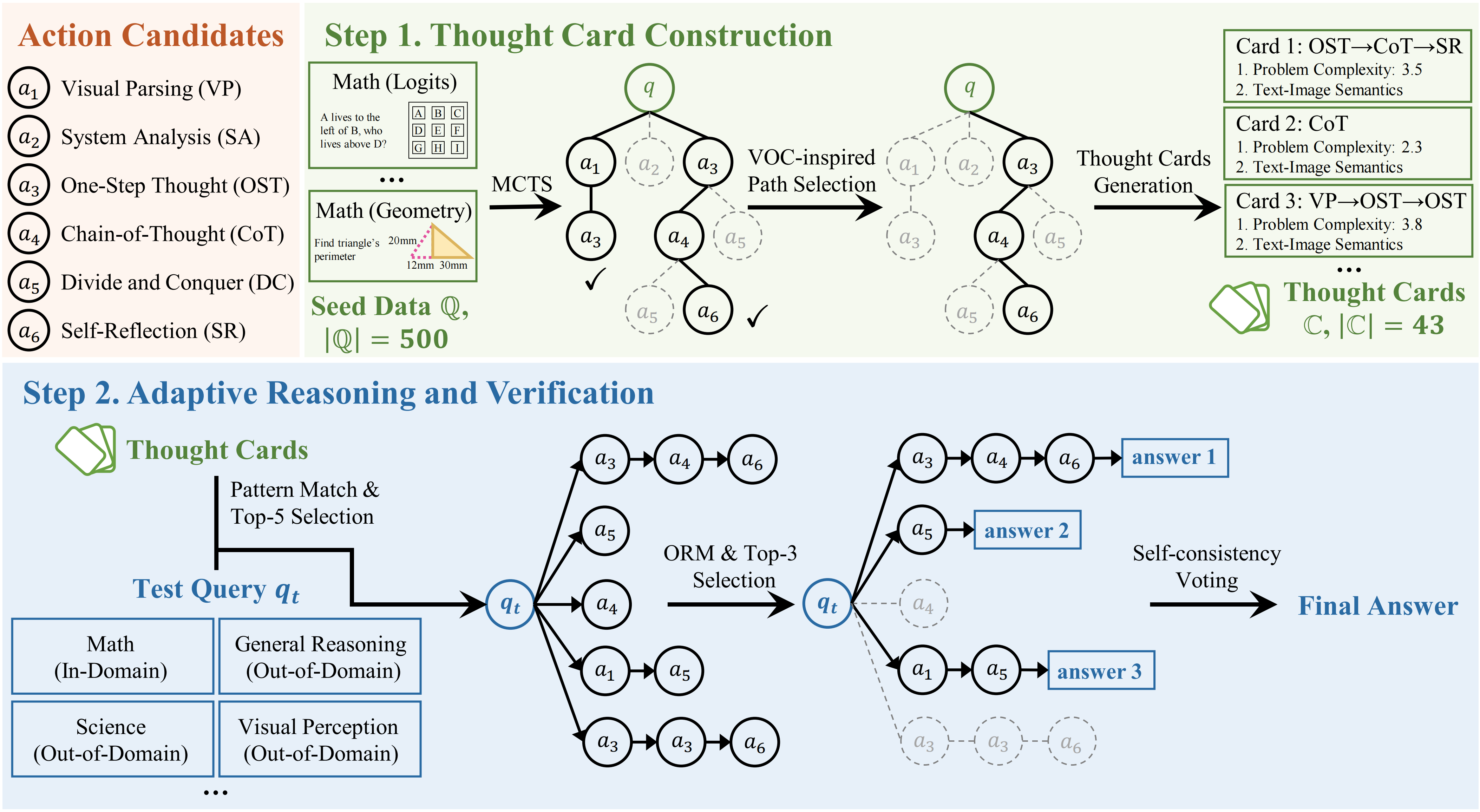}}
\caption{\textbf{Flowchart of AStar.} Building on the action candidates, this framework consists of two key components: (1) Thought Card Construction; and (2) Adaptive Reasoning and Verification.} 
\label{Figure3}
\end{center}
\end{figure*}

However, both methods face some limitations. (i) Search-based methods~\citep{dong2024progressive} suffer from computational inefficiency due to extensive exploration across solution spaces; (ii) SFT-based post-training methods~\citep{luo2025ursa,yao2024mulberry,hu2024visual} typically require substantial training data ($\ge$100K) and computational resources to implicitly extract reasoning patterns, leading to inefficient data utilization. Their reliance on proprietary models like GPT-4o for data synthesis also creates accessibility barriers for researchers outside major enterprises; (iii) RL-based post-training methods~\citep{peng2025lmm,liu2025othink,zhang2025r1} encounter challenges with exploration depth and training stability. As shown in~\citet{yue2025does}, these methods primarily function by biasing the model's output distribution toward reward-maximizing paths without introducing external knowledge. This inherently limits their exploration capacity, resulting in a narrower reasoning capability boundary compared to base models.

To address these limitations, we propose \textbf{AStar}, an automated structured thinking paradigm for multimodal reasoning. Specifically, we introduce a novel module called ``thought cards'', a lightweight, generalizable library that stores abstract, high-level reasoning patterns. These patterns are derived from small-scale prior samples using MCTS and can be effectively generalized across different domains. For each test problem, we design an adaptive retrieval mechanism to select the five most suitable thought cards based on the problem's characteristics. Leveraging these guidelines, AStar performs visual reasoning and derives the final solution through unified self-consistency checks and outcome reward scoring. Compared to existing methods, AStar eliminates the computational inefficiency of search-based approaches and removes the need for large-scale training data required by post-training methods, enabling a data- and computation-efficient reasoning framework.

Our method offers four key advantages. (i) \textbf{Performance Enhancement}: By leveraging high-level thought cards, we adaptively obtain reasoning guidelines based on problem attributes, achieving remarkable performance improvement (Figure \ref{Figure1}); (ii) \textbf{Efficiency Improvement}: Our training-free framework achieves comparable performance to powerful baselines with only small-scale prior samples, enhancing both data and computational efficiency; (iii) \textbf{Flexibility}: AStar serves as a plug-and-play test-time inference method that seamlessly integrates with other post-training techniques like SFT and PPO; (iv) \textbf{Transferability}: The thought cards generated from mathematical domains exhibit strong transferability to other reasoning tasks and also benefit general visual perception and understanding. We summarize our core contributions as follows:

\begin{itemize}

    \item We propose AStar, a training-free reasoning framework that integrates MLLMs' internal implicit reasoning capabilities with external explicit guidelines. To achieve this, we introduce ``thought cards'', a lightweight mechanism for storing high-level reasoning patterns, which can be adaptively instantiated for specific test problems.

    \item Extensive experiments demonstrate AStar's effectiveness and transferability. AStar with a 7B backbone outperforms large-scale models like GPT-4o across various benchmarks. Notably, thought cards from mathematical reasoning exhibit remarkable generalization to scientific reasoning, visual perception, and understanding tasks.

    \item AStar is a plug-and-play solution that can be integrated with popular post-training techniques. It serves as a valuable complement to existing multimodal reasoning methods, offering enhanced flexibility and scalability.

\end{itemize}

\section{Methodology}\label{sec4}
\textbf{Overview of AStar.}
\; This section introduces our proposed method AStar in detail. As shown in Figure \ref{Figure3} and \cref{algo:AStar}, our approach consists of two key steps:
\begin{itemize}
    \item \textit{Thought Card Construction}: Leverage MCTS to systematically construct thought cards, which serve as reference insights and effectively guide subsequent problem-solving during inference.
    \item \textit{Adaptive Reasoning and Verification}: Dynamically select and execute optimal reasoning patterns based on problem complexity, followed by robust solution verification.
\end{itemize}

\paragraph{Visual Reasoning Action Definition.}
% \subsection{Visual Reasoning Action Definition}\label{sec4.1}
Understanding human complex reasoning is crucial for modeling cognitive processes~\citep{Jaffe23}. Existing studies distinguish between two cognitive systems: System 1 and System 2~\citep{da2023system}. While System 1 represents fast, intuitive, yet error-prone thinking, System 2 involves slow, deliberative thinking with superior performance. With the emergence of advanced models like OpenAI o1~\citep{o1}, developing efficient System 2 to emulate human cognitive processes has gained significant research attention~\citep{yao2024mulberry,thawakar2025llamav,yang2024buffer}. Inspired by this, we introduce six vision-language reasoning actions to bridge the gap between model reasoning and human cognition: \textit{Visual Parsing (VP, $a_{1}$)}, \textit{System Analysis (SA, $a_{2}$)}, \textit{One-Step Thought (OST, $a_{3}$)}, \textit{Chain-of-Thought (CoT, $a_{4}$)}, \textit{Divide and Conquer (DC, $a_{5}$)}, \textit{Self-Reflection (SR, $a_{6}$)}. Detailed descriptions and the rationale for selecting these atomic actions are provided in Appendix C.1.

\subsection{Thought Card Construction}\label{sec4.2}
Based on the above action definitions, we introduce ``\textit{thought cards}'' as reasoning templates to guide inference in Section \ref{sec4.3}. Using only 500 prior samples, we derive their reasoning paths (Phase 1) and then distill them into high-level thought cards (Phase 2). These cards provide structured guidance for efficient problem-adaptive reasoning during inference.

\begin{algorithm}[t!]
\caption{AStar Algorithm}
\label{algo:AStar}
\textbf{Input}: a policy model $\pi_{\theta}$; a multimodal test query $q_t$; a set of seed data $\mathbb{Q}$\\
\textbf{Output}: the optimal reasoning trajectory $y_{t}$

\begin{algorithmic}[1] %[1] enables line numbers
    \STATE Initialize action space $A = \{ a_1, a_2, a_3, a_4, a_5, a_6 \}$
    \STATE Initialize repository $D \leftarrow [\;]$; Cards $\mathbb{C}\leftarrow \{\}$

    \STATE // \textit{2.1. Thought Card Construction}
    \FOR{$q \in \mathbb{Q}$}
        \STATE Acquire reasoning paths $\{q, P\} \leftarrow \text{MCTS}(\pi_{\theta};q)$
        \IF{found at least one valid reasoning path}
            \STATE Find $p_{\text{best}}$ from $P$ (Equation (\ref{eq3}))
            \STATE Add $\{q, p_{\text{best}}\}$ into $D$
            \STATE Update $\mathbb{C}$ from $D$
        \ENDIF
    \ENDFOR

    \STATE // \textit{2.2. Adaptive Reasoning and Verification}
    \STATE $\mathbb{C}_{q_t} \leftarrow \text{Card\_Match}(\mathbb{C};q_t)$
    \STATE $y_{t} \leftarrow \text{Reason\_And\_Verify}(\pi_{\theta};q_t;\mathbb{C}_{q_t})$
\end{algorithmic}
\end{algorithm}

\paragraph{Phase 1: Acquire reasoning paths for seed data.} As shown in Figure \ref{Figure3}, we employ MCTS to iteratively optimize the solution search process, generating high-quality reasoning paths for the seed dataset. This design leverages MCTS's systematic exploration and MLLMs' inherent reasoning capabilities~\citep{yin2023survey,wang2024exploring}. We formulate each multimodal reasoning problem $q \in \mathbb{Q}$ (consisting of input questions and images) as a tree search problem, where $q$ represents the root node and subsequent nodes denote reasoning steps (actions and corresponding outcomes) generated by a policy MLLM $\pi_{\theta}$. We define the state $S_{t-1}$ as the trajectory $q, s_{1},...,s_{t-1}$, where $S_{0}=q$. The next step is sampled as $s_{t}\sim \pi_{\theta}(S_{t-1})$. To guide tree expansion, we define $Q(s)$ as the reward value for node $s$. Initially, all unexplored nodes are assigned $Q(s_{i})=0$. They are updated using a weighted average between the parent's current value and its child node's value:
\begin{equation}
    Q(p) \leftarrow (1-\alpha)Q(p)+\alpha Q(s) \label{eq1}
\end{equation}
where $\alpha$ is a discount factor for future rewards. For terminal nodes, we use the likelihood of self-consistency majority vote as reward value, enabling supervision-free generalization. Specifically, this phase comprises 4 MCTS operations:

\textit{(1) Selection}. This operation identifies promising nodes for expansion. Starting from the root node, we iteratively select child nodes using the Upper Confidence Bounds applied to Trees (UCT)~\citep{11871842_29} until reaching a leaf node:
\begin{equation}
    UCT(s) = Q(s) + w\sqrt{\frac{\ln N(p) }{N(s)} }  \label{eq2}
\end{equation}
where $Q(s)$ is the reward value for node $s$, $N(s)$ is the visit count, $p$ is the parent node, and $w$ is the exploration weight. The node with the highest UCT value is selected for subsequent phases, balancing exploration and exploitation.

\textit{(2) Expansion}. The selected node $s$ is expanded by sampling $n$ actions from $\pi_{\theta}$ and generating corresponding reasoning outcomes. These $n$ child nodes are added to the tree and stored in an external memory structure.

\textit{(3) Simulation}. Starting from the selected node, we iteratively sample and expand nodes until reaching a terminal state (maximum depth or final answer node). 

\textit{(4) Backpropagation}. Upon simulation completion, node information is updated along the simulation path $s_{0},...s_{d}$. Visit counts are incremented ($N(s) \leftarrow N(s) + 1$), and node value $Q(s)$ is propagated backward to its parent node $p$ using Equation (\ref{eq1}). These updated values are used to guide subsequent UCT-based node selection.

\paragraph{Phase 2: Distill paths into thought cards.}
After executing MCTS, we obtain a tree structure for each seed dataset question, yielding multiple valid reasoning paths that constitute the path set $P$. Inspired by the concept of Value of Computation (VOC)~\citep{RUSSELL1991361}, which optimizes the trade-off between computational benefits and costs, we propose a VOC-inspired selection metric to identify the optimal reasoning trajectory from candidate solutions:
\begin{equation}
    Score(q,p_{q}) = k\cdot R(p_{q}|q) - (1-k)\cdot C(p_{q}) \label{eq3}
\end{equation}
where $q$ is the task input, $p_{q}\in P$ denotes a candidate reasoning path, and $k$ balances benefits against computational costs. Here, $R(p_{q}|q)$ denotes the path's final reward (defined as the leaf node's Q-value), while $C(p_{q})$ is the reasoning cost (defined as the number of actions in the sequence). 

Then, for each question in the seed dataset, we select the path $p_{best}$ with the highest $Score(q,p_{q})$ to build a \textit{Question-path repository} $D$ with one-to-one mappings. Inspired by metareasoning principles~\citep{RUSSELL1991361}, which advocate for adaptive reasoning strategies, we distill these question-path pairs into abstract thought cards $\mathbb{C}$ as high-level guidelines. Each card is distilled based on two attributes: problem complexity (PC)~\citep{lee2000problem} and text-image semantics (TIS)~\citep{radford2021learning}. PC represents prior known conditions derived from input image-text pairs $(i,t)$, obtained using a lightweight 2B MLLM. TIS refers to joint semantic representations based on the CLIP model encodings of $(i,t)$:
\begin{equation}
    E_q (i,t) = \frac{E_{I}(i) + E_{T}(t)}{2}
\end{equation}
Finally, each thought card contains a high-level thought template (e.g., $a_1\rightarrow a_2\rightarrow a_4$), along with the average problem complexity and text-image semantics of multiple questions sharing this template. We provide a detailed example of thought cards in Appendix C.

\subsection{Adaptive Reasoning and Verification}\label{sec4.3}
During inference, given a multimodal test query $q_t$, we compute its PC and TIS, and perform nearest neighbor matching against pre-constructed thought cards $\mathbb{C}$ to identify the five most relevant cards that best align with its complexity level and semantics. The selection process involves ranking cards based on similarity for each attribute independently:
\begin{equation} 
R_{\text{TIS}}(c) = \text{Rank}\left(E_{q_t}(i_t, t_t)^\top E_{q}(i_c, t_c)\right), \forall c \in \mathbb{C} 
\end{equation}
\begin{equation} 
R_{\text{PC}}(c) = \text{Rank}\left(\frac{1}{|PC_{q_t} - PC_{c}|}\right), \forall c \in \mathbb{C} 
\end{equation}
where $\text{Rank}(\cdot)$ assigns rankings in descending order, with higher values receiving better rankings (e.g., the highest value is ranked 1, the lowest value is ranked $|\mathbb{C}|$). We then combine these rankings to compute a total ranking score:
\begin{equation}
R(q_t, c) = R_{\text{TIS}}(c) + R_{\text{PC}}(c)
\end{equation}
Finally, we select the five thought cards with better combined rankings:
\begin{equation}
    NN_5(q_t, \mathbb{C}) = \mathop{\arg\min}_{\mathbb{C}_{q_t} \subseteq \mathbb{C}, |\mathbb{C}_{q_t}| = 5}  {\textstyle \sum_{c \in \mathbb{C}_{q_t}}}R(q_t, c) \label{eq33}
\end{equation}
where $\mathbb{C}_{q_t} \subseteq \mathbb{C}$ contains the five thought cards with the best combined rankings. We instantiate these templates for the test query to obtain five candidate solutions. To identify the best reasoning trajectory, we employ both self-consistency checks and text-domain outcome reward models due to the scarcity of visual-domain verification models. Details are described in Appendix B.4.

In summary, our approach adaptively selects high-level reasoning guidelines based on problem attributes, seamlessly integrating the model's internal implicit reasoning with external explicit guidance. This dynamic selection mechanism enhances flexibility and efficiency while maintaining robust performance across diverse problem types.

\section{Experiments}\label{sec5}

In this section, we first describe our experimental setup in detail. We then evaluate the effectiveness of AStar across four key aspects: (1) performance enhancement, (2) efficiency improvement, (3) flexibility, and (4) transferability. Finally, we conduct extensive ablation studies and analyze the impact of seed dataset size.

\subsection{Experimental Setup}\label{sec5.1}
\paragraph{Datasets.}
We conduct extensive experiments across various 4 domains and 8 datasets: (1) mathematical reasoning: MathVista~\citep{lu2023mathvista}, MathVerse~\citep{zhang2025mathverse}, and MathVision~\citep{wang2024measuring}; (2) general reasoning: MMMU~\citep{yue2023mmmu}; (3) domain-specific scientific reasoning: GAOKAO-MM~\citep{zong-qiu-2024-gaokao}; (4) visual perception and understanding: ChartQA~\citep{masry-etal-2022-chartqa}, MMStar~\citep{chen2024we}, and BLINK~\citep{10.1007/978-3-031-73337-6_9}. For all benchmarks, we use the official evaluation metrics. Further details are provided in Appendix D.1.

\paragraph{Models.}
To demonstrate AStar's versatility, we evaluate its effectiveness on both LLM and MLLM, including Qwen2.5-7B~\citep{qwen25}, and Qwen2-VL-2/7B~\citep{bai2025qwen2}. This design aims to validate that AStar can seamlessly leverage pre-trained LLMs and MLLMs as its inference backbone without modifications.

\paragraph{Baselines.}
We evaluate AStar against four representative baseline categories: (1) open-source general MLLMs like Qwen2-VL~\citep{wang2024qwen2} and InternVL2 series~\citep{chen2024internvl}; (2) open-source reasoning MLLMs, including SFT-based methods URSA~\citep{luo2025ursa} and Math-LLaVA~\citep{shi2024math}, and RL-based methods R1-VL-7B~\citep{zhang2025r1}, LMM-R1~\citep{peng2025lmm}, and MM-Eureka~\citep{meng2025mm}; (3) search-based methods like Mulberry~\citep{yao2024mulberry}; and (4) closed-source MLLMs like GPT-4o~\citep{openai2024gpt4o}.

\begin{table*}[!t]
\setlength{\tabcolsep}{1.0mm}
\centering
\begin{tabular}{l|cccccc|ccccc|cc}
\toprule
Model & \multicolumn{6}{c}{MathVerse} & \multicolumn{5}{c}{MathVista} & \multicolumn{2}{c}{MathVision} \\
\midrule
 & \textbf{ALL} & VI & VD & VO & TD & TO & \textbf{ALL} & ARI & LOG & STA & VQA & \textbf{ALL} & LOG \\
\midrule
Random & 12.4 & 12.4 & 12.4 & 12.4 & 12.4 & 12.4 & 17.9 & 13.8 & 13.4 & 14.3 & 26.3 & 7.2 & 7.6 \\
Human & 64.9 & 61.4 & 68.3 & 66.7 & 71.2 & 41.7 & 60.3 & 59.2 & 40.7 & 63.9 & 55.9 & 68.8 & 61.3 \\
GPT-4o & 50.2 & 39.6 & 42.5 & 39.3 & 54.7 & 50.0 & 60.1 & 58.4 & 27.0 & 69.0 & 49.2 & 30.4 & 29.4 \\
\midrule
\multicolumn{14}{c}{\textit{Open-Source General MLLMs}} \\
\midrule
MiniGPT4-7B~\cite{zhu2023minigpt} & 12.2 & 12.5 & 14.8 & 8.7 & 12.3 & 13.4 & 23.1 & 32.0 & 10.8 & 17.9 & 30.2 & 10.8 & 12.7 \\
LLaVA-1.5-13B~\cite{liu2024improved} & 12.7 & 12.6 & 12.7 & 9.0 & 17.1 & 22.6 & 27.7 & 28.6 & 10.8 & 22.9 & 30.2 & 11.1 & 13.5 \\
SPHINX-V2-13B~\cite{lin2023sphinx} & 16.1 & 16.4 & 15.6 & 16.2 & 20.8 & 14.0 & 36.7 & 33.4 & 24.3 & 51.5 & 43.0 & 9.7 & 10.1 \\
SPHINX-8x7B~\cite{lin2023sphinx} & 22.8 & 21.1 & 19.6 & 18.3 & 33.3 & 23.1 & 42.6 & 43.0 & 14.4 & 50.8 & 43.3 & 15.8 & 17.9 \\
LLaVA-NeXT-34B~\cite{liu2024llava} & 34.6 & 35.2 & 28.9 & 22.4 & 49.0 & 30.1 & 46.5 & - & - & - & - & - \\
InternVL2-8B~\cite{chen2024internvl} & 35.9 & 32.2 & 30.9 & 27.7 & 39.0 & 36.0 & 58.3 & 56.4 & 10.8 & 68.8 & 49.7 & 18.4 & 15.3 \\
Qwen2-VL-7B~\cite{wang2024qwen2} & 33.6 & 31.3 & 30.3 & 28.1 & 37.4 & 35.0 & 58.9 & 57.5 & 24.3 & 43.1 & 58.1 & 17.2 & 12.7 \\
\midrule
\multicolumn{14}{c}{\textit{Open-Source Reasoning MLLMs (Large-Scale Post-Training)}} \\
\midrule
Math-LLaVA-13B~\cite{shi2024math} & 22.9 & 24.5 & 21.7 & 16.1 & 27.3 & 27.0 & 46.6 & 40.7 & 23.3 & 42.3 & 33.5 & 15.7 & 16.0 \\
Math-PUMA-7B~\cite{zhuang2024math} & 33.6 & 33.4 & 31.6 & 26.0 & 42.1 & 39.8 & 47.9 & 46.2 & 21.6 & 55.8 & 30.2 & 14.0 & 13.7 \\
MultiMath-7B~\cite{peng2024multimath} & 27.7 & 28.1 & 25.9 & 15.0 & 34.8 & 35.3 & 50.0 & 42.2 & 23.3 & 64.9 & 49.2 & 16.3 & 17.9 \\
URSA-8B~\cite{luo2025ursa} & 45.7 & 46.4 & 43.9 & 28.6 & 55.3 & 51.8 & 59.8 & 53.5 & 21.6 & 57.1 & 40.2 & 26.2 & 24.8 \\
R1-VL-7B~\cite{zhang2025r1} & 40.0 & 37.3 & 33.6 & 39.8 & 45.0 & 40.7 & 63.5 & 54.7 & 28.5 & 61.2 & \textbf{60.9} & 27.1 & 23.6 \\
\midrule
AStar (Qwen2.5-7B, Training-free) & \textbf{53.9} & \textbf{49.7} & \textbf{64.4} & \textbf{48.6} & \textbf{56.4} & \textbf{56.1} & \textbf{64.2} & \textbf{63.8} & \textbf{59.5} & \textbf{69.1} & \textbf{60.9} & \textbf{32.7} & \textbf{39.4} \\
AStar (Qwen2-VL-7B, Training-free) & 47.5 & 41.8 & 51.6 & \textbf{48.6} & 49.3 & 42.2 & 61.7 & 60.6 & 28.6 & 68.4 & 59.6 & 27.9 & 29.4 \\
\bottomrule
\end{tabular}
\caption{\textbf{Performance comparison on MathVista, MathVerse, and MathVision.} The best model results are highlighted in \textbf{bold}. For MathVerse, we show 6 categories: ALL (overall accuracy), VI (vision intensive), VD (vision dominant), VO (vision only), TD (text dominant), and TO (text only). For MathVista, we present 5 categories: ALL (overall accuracy), ARI (arithmetic reasoning), LOG (logical reasoning), STA (statistical reasoning), and VQA (visual question answering). For MathVision, we present 2 categories: ALL (overall accuracy) and LOG (logical reasoning).}
\label{table1}
\end{table*}

\subsection{Performance Enhancement}\label{sec3.1 performance enhancement}
\paragraph{Main Results.}
Table \ref{table1} presents the performance of AStar across three representative multimodal reasoning benchmarks. We have four key findings:

$\circ$ AStar consistently outperforms both general and math-specialized MLLMs. With the Qwen2.5-7B reasoning backbone, AStar achieves $53.9\%$ accuracy on MathVerse, surpassing large-scale CoT-trained URSA-8B by $8.2\%$, and expensive GRPO-trained R1-VL-7B by $13.9\%$.

$\circ$ AStar demonstrates strong performance across diverse reasoning types. It reaches $59.5\%$ on logical reasoning (LOG), outperforming the powerful URSA-8B by $37.9\%$. Similar gains are observed in statistical reasoning (STA: $69.1\%$) and visual question answering (VQA: $60.9\%$).

$\circ$ AStar's adaptive reasoning benefits are universal across varying multimodal information distributions. It maintains consistent gains from vision-dominant (VD: 64.4$\%$) to text-dominant (TD: 56.4$\%$) scenarios, showcasing robust performance regardless of modality balance. Its strong performance in text-only (TO) scenarios highlights the paradigm's versatility and ability to generalize to pure text domains.

$\circ$ On the more challenging MathVision benchmark~\citep{wang2024measuring}, AStar achieves 32.7$\%$ overall accuracy, surpassing GPT-4o (30.4$\%$). Notably, in logical reasoning, AStar attains 39.4$\%$ accuracy, significantly outperforming GPT-4o (29.4$\%$) by 10.0$\%$. This stems from our adaptive decomposition framework that transforms complex reasoning into manageable sub-problems.

\paragraph{High Performance with Efficient Models.}
As shown in Figure \ref{Figure1}, we intuitively demonstrate AStar's effectiveness across different model architectures and scales. Notably, AStar (Qwen2.5-7B) achieves 53.9$\%$ accuracy on MathVerse, surpassing GPT-4o's 50.2$\%$. Even smaller models like AStar+Qwen2-VL-2B outperform larger baselines like Qwen2-VL-7B and InternVL2-40B. This demonstrates AStar's capacity to boost smaller models to competitive performance against significantly larger architectures. Detailed results are provided in Appendix E.2.

\begin{table*}[ht!]
\setlength{\tabcolsep}{0.9mm}
    \centering
    \begin{tabular}{l|cccccccc}
    \toprule  
    Methods & Type & OS Only & Training-Free & Prior Data~$\downarrow$ & Pre. Time~$\downarrow$ & MMStar~$\uparrow$ & MathVerse~$\uparrow$ & MathVision~$\uparrow$ \\
    \midrule
    Mulberry-7B & Search & \xmark & \xmark & 260K & - & 61.3 & 44.9 & 26.4 \\
    URSA-8B & SFT & \xmark & \xmark & 1100K & 3 days & 55.4 & 45.7 & 26.2 \\
    LMM-R1-3B & PPO & \cmark & \xmark & 55.3K & - & 58.0 & 41.8 & 26.9 \\
    R1-VL-7B & GRPO & \cmark & \xmark & 260K & - & 60.0 & 40.0 & 27.1 \\
    MM-Eureka-7B & GRPO & \cmark & \xmark & 15K & - & 59.4 & 50.3 & 26.9 \\
    \textbf{Ours (Qwen2.5-7B)} & - & \cmark & \cmark & \textbf{0.5K} & \textbf{50 mins} & \textbf{62.3} & \textbf{53.9} & \textbf{32.7} \\
    \textbf{Ours (Qwen2-VL-7B)} & - & \cmark & \cmark & \textbf{0.5K} & \textbf{50 mins} & \textbf{62.0} & \textbf{47.5} & \textbf{27.9} \\
    \bottomrule
  \end{tabular}
  \caption{\textbf{Efficiency comparison.} `OS Only' indicates exclusive use of open-source models. `Pre. Time' denotes preprocessing or training time, specifically thought card construction time for our method and dataset construction/training time for others.}
    \label{table3}
\end{table*}

\subsection{Efficiency Improvement}\label{sec3.2 efficiency improvement}
To demonstrate AStar's efficiency, we compare it against search-based (Mulberry) and post-training methods (URSA, LMM-R1, MM-Eureka, R1-VL). Table \ref{table3} evaluates resource requirements, accessibility, and performance on visual understanding (MMStar) and mathematical reasoning (MathVerse, MathVision) benchmarks.

AStar achieves superior performance with exceptional efficiency. Our approach requires only 0.5K prior samples and 50 minutes of preprocessing time. Our approach requires only 0.5K prior samples and 50 minutes for thought card generation (see Figure~\ref{Figure3}), eliminating the need for any training process or model parameter updates. This results in a 520-fold and 2200-fold reduction in data requirements compared to Mulberry (260K) and URSA-8B (1100K), while eliminating URSA-8B's 3-day training requirement. Despite these significant efficiency gains, AStar consistently outperforms baselines: MMStar (62.3$\%$ vs. 61.3$\%$ for Mulberry), MathVerse (53.9$\%$ vs. 45.7$\%$ for URSA-8B), and MathVision (32.7$\%$ vs. 26.9$\%$ for MM-Eureka-7B). This stems from our explicit reasoning pattern extraction that captures problem-solving strategies without extensive computational overhead. We think our framework may provide insights for resource-constrained researchers.

\subsection{Flexibility}\label{sec3.3 flexibility}
\begin{figure}
  \centering
  \includegraphics[width=0.95\linewidth]{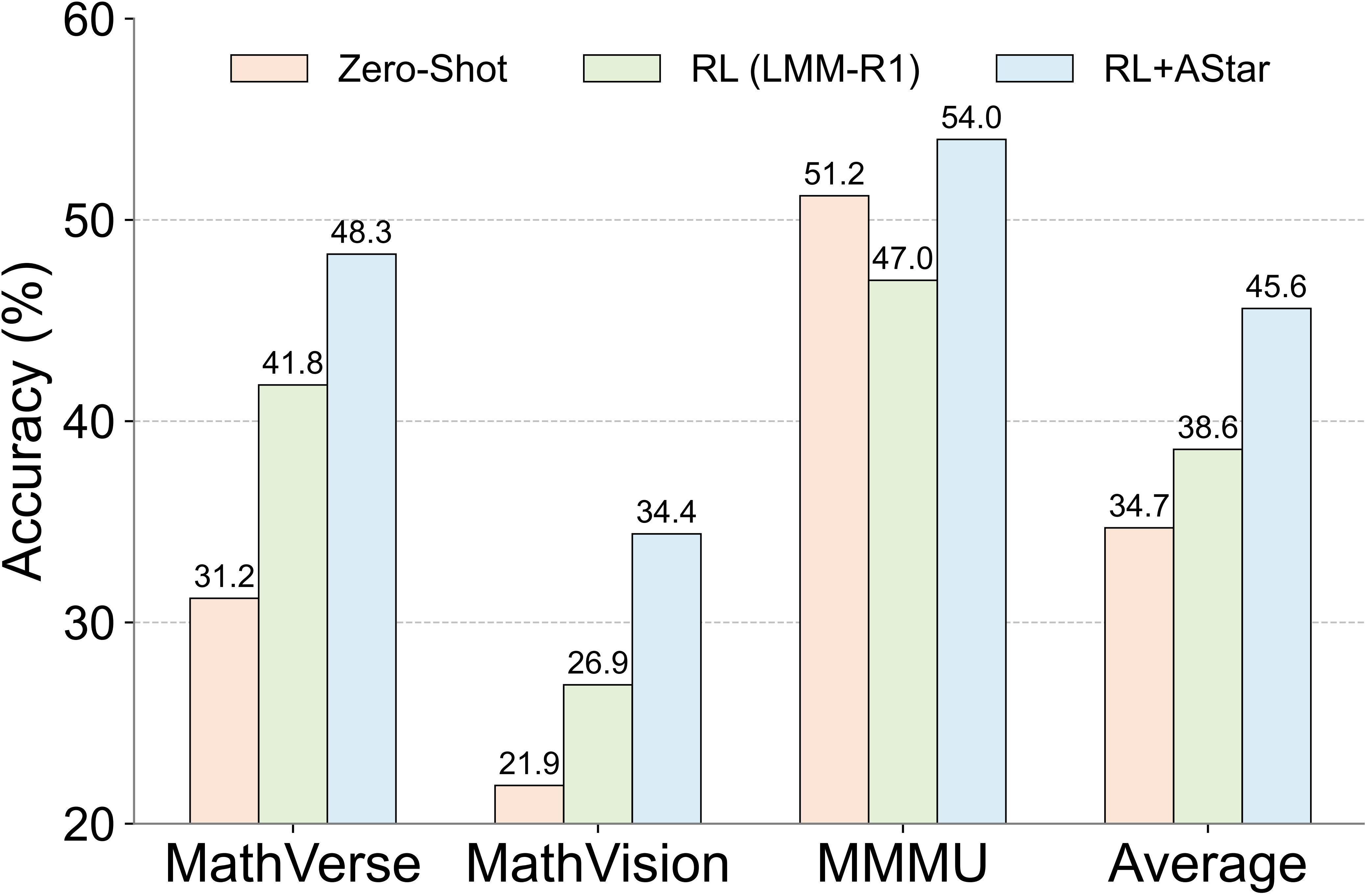}
  \caption{\textbf{Flexibility verification.} AStar is a plug-and-play framework that can be integrated with RL-trained models (from Qwen2.5-VL-3B) for further improvement.}
  \label{Figure8}
\end{figure}

As a test-time inference framework, AStar can be effectively integrated with post-training methods. We demonstrate this by applying our reasoning paradigm to PPO-trained LMM-R1~\citep{peng2025lmm} from Qwen2.5-VL-3B. Figure \ref{Figure8} shows that AStar consistently enhances performance across all benchmarks when combined with post-trained models. The combination (RL+AStar) achieves 48.3$\%$ accuracy on MathVerse, improving upon the base RL model (41.8$\%$) by 6.5$\%$. This synergy indicates AStar captures complementary reasoning patterns beyond post-training, highlighting its plug-and-play adaptability. Additional integration with SFT and GRPO are provided in the Appendix E.5.

\begin{figure*}[ht!]
\centerline{\includegraphics[width=0.98\linewidth]{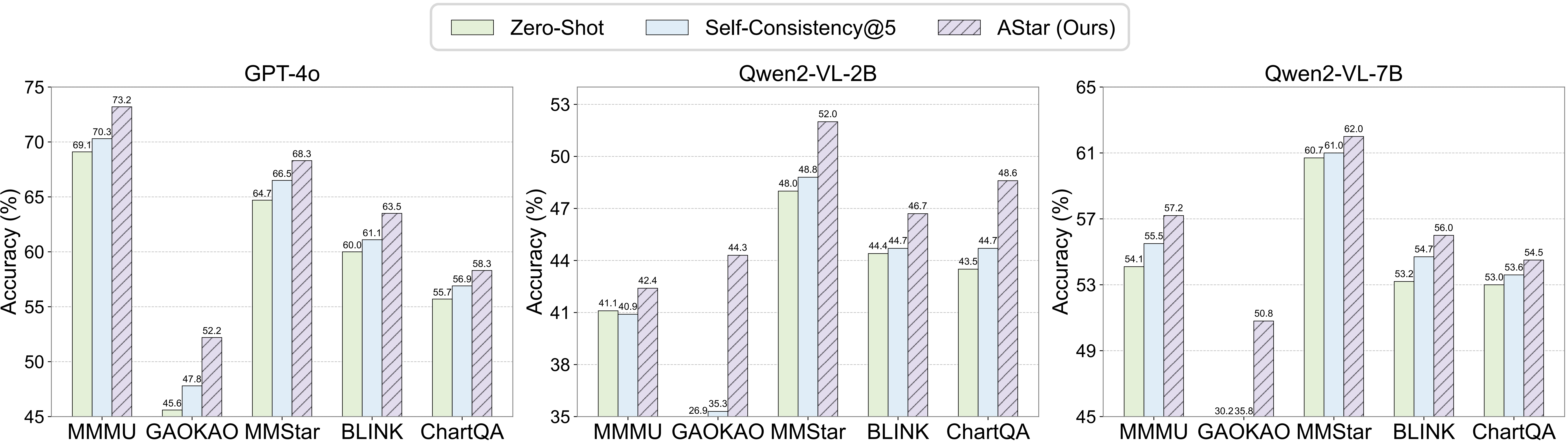}}
\caption{\textbf{Transferability verification.} Beyond math reasoning, we conduct extensive cross-domain experiments on diverse tasks, including general reasoning (MMMU), domain-specific reasoning (GAOKAO), visual perception and comprehension (MMStar and BLINK), and chart understanding (ChartQA). Results on ChartQA are consistently scaled for better visualization, which does not affect the conclusions. Our framework exhibits remarkable cross-domain generalization.}
\label{Figure6}
\end{figure*}

\begin{table}[t]
\setlength{\tabcolsep}{0.8mm}
\centering
\begin{tabular}{lccl}
\toprule
Model Setting & MathVision & MathVerse & Average \\
\midrule
AStar (Full) & \textbf{32.7} & \textbf{53.9} & \textbf{43.3} \\
\midrule
\;\; w/o Thought Cards & 24.8 & 42.9 & 33.8$^{\textbf{\scriptsize-9.5}\downarrow}$ \\
\;\; w/o Card Matching & 27.9 & 47.2 & 37.6$^{\textbf{\scriptsize-5.7}\downarrow}$ \\
\;\; w/ Random Selection & 30.9 & 51.7 & 41.3$^{\textbf{\scriptsize-2.0}\downarrow}$ \\
\;\; w/ Self-Consistency & 31.5 & 52.0 & 41.8$^{\textbf{\scriptsize-1.5}\downarrow}$ \\
\bottomrule
\end{tabular}
\caption{\textbf{Ablation study.} Using Qwen2.5-7B as the backbone, we systematically evaluate the impact of: (1) removing thought cards (replaced with random action combinations), (2) disabling card matching (replaced with random cards), and (3) replacing verification with simpler alternatives (replaced with random selection or self-consistency).}
\label{table4}
\end{table}

\subsection{Transferability}\label{sec3.4 transferability}

Recent work has highlighted that distributional shifts severely affect MLLM reliability~\citep{yin2023survey,wang2024exploring}. While these models excel on in-domain tasks, their performance often degrades in out-of-domain (OOD) scenarios~\citep{dong2024multiood}. This challenge is compounded by the difficulty of acquiring sufficient domain-specific training data and computational resources.

To evaluate AStar's cross-domain transferability, we investigate whether thought cards constructed from mathematical domains can generalize to diverse non-mathematical tasks. Since our thought cards derive from math domains (details in Appendix D.3), we conduct OOD evaluations on general reasoning (MMMU), domain-specific reasoning (GAOKAO), visual perception (MMStar, BLINK), and chart understanding (ChartQA). Figure \ref{Figure6} shows AStar's consistent improvements across all non-mathematical domains. For instance, AStar with Qwen2-VL-2B achieves notable gains on GAOKAO (44.3$\%$ vs. 35.3$\%$ SC@5), and MMStar (52.0$\%$ vs. 48.8$\%$), while Qwen2-VL-7B shows similar improvements. Remarkably, even for the powerful GPT-4o, AStar provides consistent enhancements across benchmarks like MMMU (73.2$\%$ vs. 70.3$\%$) and GAOKAO-MM (52.2$\%$ vs. 47.8$\%$). This demonstrates that mathematical thought cards can successfully transfer to diverse domains, validating that high-level abstract reasoning patterns provide robust cross-domain generalization. We further explore weak-to-strong generalization in Appendix E.4, where reasoning guidance from smaller models (Qwen2-VL-7B) even enhances GPT-4o's performance, supporting our framework's universal applicability.

\subsection{Ablation Study and Analysis}\label{sec3.6 ablation study}
\paragraph{Ablation Study.}
As shown in Table \ref{table4}, we remove or replace each key component in AStar to understand its individual performance impacts:

$\circ$ \textit{Structured reasoning patterns matter.} Replacing thought cards with random action sequences results in a 9.5$\%$ performance drop, highlighting the importance of systematically extracted reasoning patterns.

$\circ$ \textit{Problem-pattern matching is crucial.} Different matching metrics yield varying performance, demonstrating that retrieving appropriate thought cards based on problem characteristics is vital for optimal performance. This validates our adaptive selection mechanism.

$\circ$ \textit{Verification provides consistent gains.} While replacing our verification module with simpler alternatives shows modest degradation (e.g., $1.5\%$ on self-consistency), the results indicate that thought cards enable robust solution generation even with basic verification strategies.

\paragraph{Impact of Seed Dataset Size.}
As shown in Table~\ref{table_sensitivity}, the performance varies with the size of the seed dataset. On average, the accuracy improves from 33.5\% (with 50 samples) to 44.1\% (with 1,000 samples). Notably, AStar achieves competitive accuracy even with just 100 samples, highlighting its strong practicality in resource-constrained settings. In this paper, we select 500 seed samples by default, balancing performance and computational efficiency.

\begin{table}[t!]
\centering
% The column specifier was changed to cccc for four columns
\begin{tabular}{cccc} 
    \toprule
    Seed Data & MathVision~$\uparrow$ & MathVerse~$\uparrow$ & Average~$\uparrow$ \\
    \midrule
    50 & 23.8 & 43.2 & 33.5 \\
    100 & 28.9 & 49.8 & 39.4 \\
    200 & 30.8 & 51.8 & 41.3 \\
    500 & 32.7 & 53.9 & 43.3 \\
    1000 & 33.4 & 54.8 & 44.1 \\
    \bottomrule
\end{tabular}
\caption{Impact of seed dataset size}
\label{table_sensitivity}
\end{table}

\section{Related Work}\label{sec2}
\paragraph{Multimodal Reasoning.}
MLLMs have demonstrated robust capabilities across diverse domains~\citep{zhang2024mavis,du2025virgo}. However, complex multimodal reasoning remains challenging, requiring both visual perception and high-level cognitive capacity. While recent methods~\citep{xu2024llava,thawakar2025llamav} implement structured reasoning to enhance CoT capabilities~\citep{zhang2024multimodal}, they face critical limitations: (i) explicit search methods~\citep{dong2024progressive,yao2024mulberry} suffer from computational inefficiency, and (ii) post-training methods~\citep{wang2025multimodal,zhang2025r1} demand substantial resources. These methods rely on SFT with limited generalization or RL techniques like GRPO that lack external knowledge integration~\citep{yue2025does,wu2025thought,yang2025reasonflux}. Our approach leverages high-level thought cards to seamlessly integrate internal model capabilities with external reasoning guidelines. Additionally, existing methods typically employ rigid structures that limit flexibility, overlooking the importance of adaptive reasoning in unleashing reasoning potential~\citep{wang2024enhancing}. Our thought cards enable task-specific reasoning path generation, balancing performance with efficiency.

\paragraph{Tree-based Search.}
Tree structures have shown significant potential in language models~\citep{qi2024mutual,wu2024beyond}. Recent research applies them to identify effective reasoning paths for MLLMs. AR-MCTS~\citep{dong2024progressive} integrates MCTS with active retrieval but suffers from extensive iterations. Mulberry~\citep{yao2024mulberry} distills 260K reasoning chains from GPT-4o using tree structures, yet they require substantial computational resources and high-capacity teacher models. These methods fail to balance performance with efficiency. We incorporate high-level problem-solving guidelines into MCTS, achieving competitive performance with higher efficiency. While using MCTS as our reasoning organization structure, our framework also accommodates other advanced structures in future work.

\section{Conclusion}\label{sec6}
We introduce AStar, a training-free structured thinking paradigm for multimodal reasoning. By leveraging \textit{thought cards} and adaptively retrieving problem-specific guidelines, our method effectively integrates MLLMs' internal capabilities with external explicit guidance. AStar-7B achieves 53.9$\%$ accuracy on MathVerse and 32.7$\%$ on MathVision, surpassing GPT-4o. Further analysis reveals that thought cards from math domains also benefit other tasks like visual perception. As a plug-and-play test-time framework, AStar offers a promising pathway for multimodal reasoning.

\section{Acknowledgments}
This work is supported by the National Natural Science Foundation of China (No. U2436210), Excellent Youth Program of State Key Laboratory of Multimodal Artificial Intelligence Systems (No. MAIS2024311) and  Youth Science Fund Project of National Natural Science Foundation of China (No. 62201572).

\bibliography{aaai2026}

% \clearpage

% \input{appendix}

\end{document}